\documentclass[nohyperref]{article}

\usepackage{microtype}
\usepackage{graphicx}
\usepackage{booktabs}
\usepackage{hyperref}

\usepackage[accepted]{icml2024}

\usepackage{amsmath}
\usepackage{amssymb}
\usepackage{mathtools}
\usepackage{amsthm}
\usepackage{amsfonts}       
\usepackage{nicefrac}       
\usepackage{microtype}      
\usepackage{xcolor}         
\usepackage{xspace}
\usepackage{colortbl}
\usepackage{graphicx}
\usepackage{enumitem}
\usepackage{subcaption}
\usepackage{epigraph}
\usepackage{changepage}
\usepackage{multirow}
\usepackage{booktabs} 
\usepackage{pifont}
\usepackage{adjustbox}
\usepackage{lipsum} 
\usepackage{duckuments}

\usepackage{algorithm}
\usepackage{listings}

\usepackage{amsmath,amsfonts,bm}

\def\eqref#1{equation~\ref{#1}}

\def\1{\bm{1}}

\DeclareMathAlphabet{\mathsfit}{\encodingdefault}{\sfdefault}{m}{sl}
\SetMathAlphabet{\mathsfit}{bold}{\encodingdefault}{\sfdefault}{bx}{n}

\def\gF{{\mathcal{F}}}

\def\gJ{{\mathcal{J}}}

\newcommand{\blap}[1]{\vbox to 0pt{\hbox{#1}\vss}}

\newlength\savewidth\newcommand\shline{\noalign{\global\savewidth\arrayrulewidth
  \global\arrayrulewidth 1pt}\hline\noalign{\global\arrayrulewidth\savewidth}}
\newcommand{\tablestyle}[2]{\setlength{\tabcolsep}{#1}\renewcommand{\arraystretch}{#2}\centering\footnotesize}
\definecolor{baselinecolor}{HTML}{EEEEEE}
\newcommand{\baseline}[1]{\cellcolor{baselinecolor}{#1}}

\usepackage[capitalise,noabbrev,nameinlink]{cleveref}
\creflabelformat{equation}{#2\textup{#1}#3}
\creflabelformat{section}{#2\textup{#1}#3}
\creflabelformat{appendix}{#2\textup{#1}#3}
\creflabelformat{figure}{#2\textup{#1}#3}
\crefname{algocf}{Algorithm}{Algorithms}
\Crefname{algocf}{Algorithm}{Algorithms}

\definecolor{DarkBlue}{rgb}{0,0.08,0.45}

\hypersetup{%
colorlinks=true,
linkcolor=DarkBlue,
citecolor=DarkBlue,
filecolor=DarkBlue,
urlcolor=DarkBlue}

\theoremstyle{plain}

\theoremstyle{definition}

\theoremstyle{remark}

\usepackage[textsize=tiny]{todonotes}

\definecolor{JSViolet}{RGB}{71,15,244}
\definecolor{JSRed}{RGB}{205,44,78}
\definecolor{RowHighlight}{gray}{0.9}
\newcommand{\cmark}{\textcolor{JSViolet}{\ding{51}}}
\newcommand{\xmark}{\textcolor{JSRed}{\ding{55}}}

\icmltitlerunning{Visual Representation Learning with Stochastic Frame Prediction}

\begin{document}

\twocolumn[
\icmltitle{Visual Representation Learning with Stochastic Frame Prediction}

\begin{icmlauthorlist}
\icmlauthor{Huiwon Jang}{kaist}
\icmlauthor{Dongyoung Kim}{kaist}
\icmlauthor{Junsu Kim}{kaist}
\icmlauthor{Jinwoo Shin}{kaist}
\icmlauthor{Pieter Abbeel}{berkeley}
\icmlauthor{Younggyo Seo}{kaist,dyson}
\end{icmlauthorlist}

\icmlaffiliation{kaist}{KAIST}
\icmlaffiliation{berkeley}{UC Berkeley}
\icmlaffiliation{dyson}{Now at Dyson Robot Learning Lab}

\icmlcorrespondingauthor{Huiwon Jang}{huiwoen0516@kaist.ac.kr}
\icmlcorrespondingauthor{Younggyo Seo}{mail@younggyo.me}

\icmlkeywords{Machine Learning, ICML}

\vskip 0.3in
]



\printAffiliationsAndNotice{}  

\begin{abstract}
Self-supervised learning of image representations by predicting future frames is a promising direction but still remains a challenge.
This is because of the under-determined nature of frame prediction; multiple potential futures can arise from a single current frame.
To tackle this challenge, in this paper, we revisit the idea of stochastic video generation that learns to capture uncertainty in frame prediction and explore its effectiveness for representation learning.
Specifically, we design a framework that trains a stochastic frame prediction model to learn temporal information between frames.
Moreover, to learn dense information within each frame, we introduce an auxiliary masked image modeling objective along with a shared decoder architecture.
We find this architecture allows for combining both objectives in a synergistic and compute-efficient manner.
We demonstrate the effectiveness of our framework on a variety of tasks from video label propagation and vision-based robot learning domains, such as video segmentation, pose tracking, vision-based robotic locomotion, and manipulation tasks. Code is available on the project webpage: \href{https://sites.google.com/view/2024rsp}{https://sites.google.com/view/2024rsp}.
\end{abstract}

\begin{figure*}[tb]
\centering
\begin{subfigure}{0.615\textwidth}
  \centering
  \includegraphics[width=0.95\linewidth]{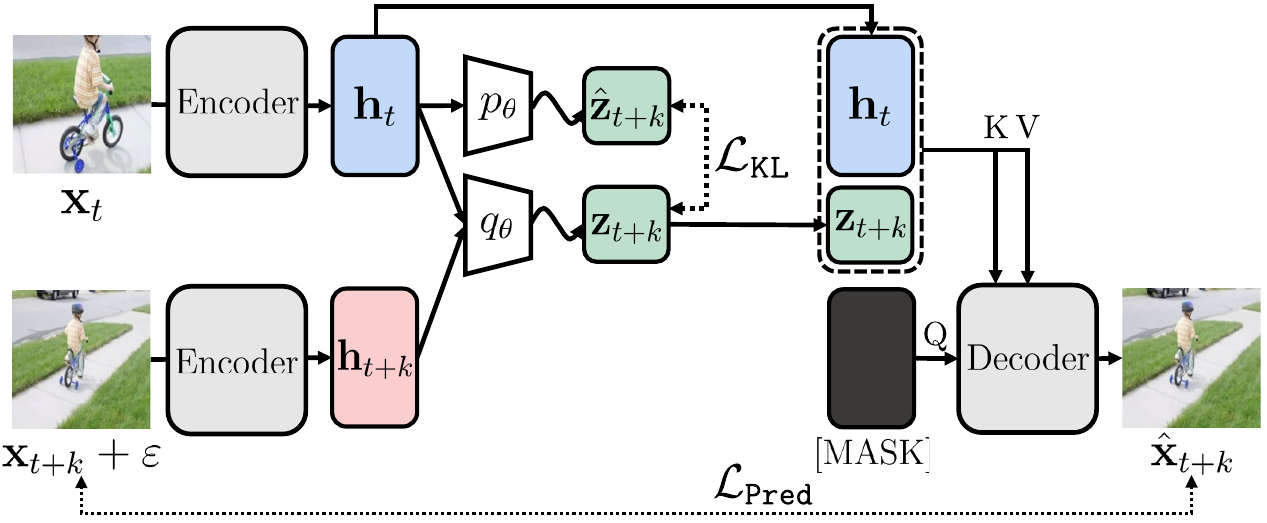}
  \caption{Stochastic frame prediction}
  \label{fig:method_sub1}
\end{subfigure}%
\begin{subfigure}{0.375\textwidth}
  \centering
  \includegraphics[width=0.95\linewidth]{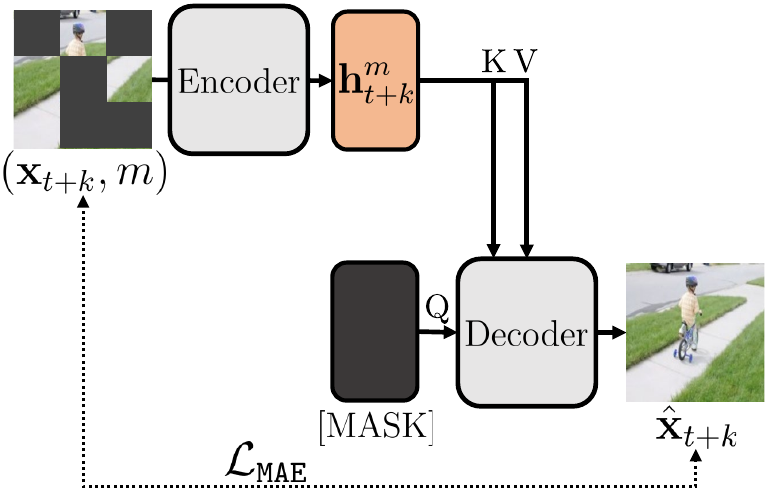}
  \caption{Masked autoencoding with shared decoder}
  \label{fig:method_sub2}
\end{subfigure}
\vspace{-0.075in}
\caption{\textbf{Representation learning with stochastic frame prediction.} (a) We train a stochastic frame prediction model, which is built upon stochastic video generation model \citep{denton2018stochastic}, which consists of an encoder that extracts representations, a posterior model with access to both current and future frames, a prior model with only access to the current frame, and a decoder that generates frame conditioned on features from the current frame and a sample from either posterior or prior distributions. We train the model to accurately generate the future frame while enforcing the posterior and prior distributions to be close to each other, \textit{i.e.,} encourage the posterior distribution to be more predictable and the prior distribution to predict the future. (b) We introduce an auxiliary masked autoencoding objective \citep{he2022masked} with a shared decoder architecture. Our decoder makes the \texttt{[MASK]} tokens attend to different inputs via the cross-attention layer, enabling us to share the decoder parameters for different objectives.}
\label{fig:method_overview}
\vspace{-0.0625in}
\end{figure*}
\section{Introduction}
Recently, generative pre-training on sequential data has been extremely successful in learning models that can be easily fine-tuned \citep{oord2016wavenet,yang2019xlnet,dai2019transformer,radford2019language} or achieve impressive performance with few adaptations or even without adaptation \citep{brown2020language,touvron2023llama}.
The core idea behind these successes is training the model to \textit{predict the future}, \textit{i.e.,} learning the distribution of future outputs conditioned on past data consisting of words \citep{bengio2000neural,radford2019language}, audio signals \citep{oord2016wavenet,dhariwal2020jukebox}, or state of the world \citep{chen2021decision}, enabling the models to understand the temporal and causal relationships within the data.

There also have been efforts to learn rich representations in video domains by learning video prediction models \citep{srivastava2015unsupervised,vondrick2016generating,finn2016unsupervised,yu2020efficient}, for its promise of utilizing an abundance of videos for learning representations that understand how the world operates by predicting the future.
However, it has been less successful when compared to its counterparts in image domains \citep{kingma2013auto,donahue2019large,chen2020generative,li2023mage} or other self-supervised learning approaches that do not involve generative modeling of future frame \citep{wang2015unsupervised,misra2016shuffle,sermanet2018time,han2020self}.

In this paper, we argue that this challenge can be attributed to the inherently under-determined nature of future frame prediction, where multiple potential futures can arise from a single current frame \citep{babaeizadeh2017stochastic,denton2018stochastic}.
This issue makes it difficult for deterministic models
to learn useful representations from complex real-world videos because the model would struggle to approximate the multi-modal distribution of future frames.
In contrast, recent video generation models have achieved remarkable successes in generating high-fidelity videos \citep{yan2021videogpt,villegas2022phenaki,ho2022imagen,blattmann2023stable,blattmann2023align},
where the core idea is to train a \textit{stochastic} generative model\footnote{
While a deterministic prediction model learns a deterministic mapping from the current frame to the future frame, a stochastic prediction model aims to learn a distribution over the future frame conditioned on the current frame.}
that can capture the uncertainty in generating or predicting the videos, such as denoising diffusion models \citep{ho2022imagen,yu2023video} and autoregressive models \citep{yan2021videogpt,villegas2022phenaki}.
Inspired by these successes, we aim to investigate how to adopt and utilize the idea of training a stochastic generative model for visual representation learning from videos.

\paragraph{Contribution} We present visual Representation learning with Stochastic frame Prediction (RSP), a framework for visual representation learning from videos.
Our key idea is to learn \textit{image} representations that capture temporal information between frames by learning a stochastic frame prediction model with videos.
To this end, we revisit the idea of stochastic video generation \citep{denton2018stochastic} that trains a time-dependent prior over future frames to capture uncertainty in frame prediction (see \cref{fig:method_sub1}).
Specifically, our key contribution lies in exploring various design choices and incorporating recent advances in self-supervised learning into the video generation model \citep{dosovitskiy2020image,hafner2020mastering,gupta2023siamese}, to re-configure it for representation learning.
We find that RSP allows for learning strong image representations from complex real-world videos when compared to deterministic prediction objectives.
To learn dense information within each frame, we further introduce an auxiliary masked autoencoding objective \citep{he2022masked}, along with a shared decoder architecture that enables us to incorporate the auxiliary objective in a synergistic manner (see \cref{fig:method_sub2}). 

Through extensive experiments, we show that RSP can effectively learn image representations from a large real-world video dataset.
Pre-trained on Kinetics-400 dataset \citep{kay2017kinetics},
RSP achieves competitive or superior performance to various self-supervised learning baselines on a variety of tasks from vision-based robot learning benchmarks \citep{james2020rlbench,majumdar2023we} and video label propagation benchmarks \citep{pont20172017,zhou2018adaptive,jhuang2013towards}.
In particular, RSP achieves a 36.0\% average success rate in challenging robotic manipulation tasks from RLBench \citep{james2020rlbench}, while MAE baseline only achieves a 13.5\% success rate.
We also provide extensive ablation studies and analyses on the importance of various design choices in our framework.

\section{Related Work}
\paragraph{Image self-supervised learning}
Self-supervised learning (SSL) from images has demonstrated remarkable success in visual representation learning by exploiting the rich, inherent structure of visual data without human labels \citep{chen2020simple, he2020momentum,  chen2021empirical, caron2021emerging, he2022masked}.
Pioneer works for SSL propose pretext tasks \citep{doersch2015unsupervised, pathak2016context, zhang2016colorful, noroozi2016unsupervised, gidaris2018unsupervised}, and recently, contrastive learning \citep{chen2020simple, he2020momentum, chen2021empirical, caron2021emerging} and masked image modeling \citep{bao2021beit, he2022masked, xie2022simmim, li2023mage} have gained prominence.
In this paper, we show that integrating an understanding of the temporal information between the frames can further enhance image representation.

\paragraph{Video self-supervised learning}
Most prior researches on SSL from videos aim to learn video representations capturing spatiotemporal information from videos that could be useful for video understanding tasks such as action recognition \citep{xu2019self, benaim2020speednet, han2020memory,han2020self,feichtenhofer2021large,pan2021videomoco,qian2021spatiotemporal,ge2021revitalizing,guo2022cross,tong2022videomae,feichtenhofer2022masked}.
Our work differs in that we focus on learning useful image representations from videos.
Similarly to our work, there have been approaches that focus on enhancing image representations, by designing pretext tasks for videos \citep{wang2015unsupervised, misra2016shuffle}, extending contrastive learning to video frames \citep{sermanet2018time,wang2019learning,jabri2020space,xu2021rethinking}, and masked visual modeling  \citep{feichtenhofer2022masked,gupta2023siamese}.
In particular, \citet{gupta2023siamese} learns visual correspondence by predicting the masked patches from the future frame.
This is closely related to our work as it represents another approach to the future frame prediction objective.
However, unlike \citet{gupta2023siamese}, which resolves ambiguity about the future by conditioning on unmasked patches from the future frame, we aim to learn representations that capture the inherent stochasticity of future frame prediction.

\section{Method}
In this section, we present \textbf{R}epresentation learning with \textbf{S}tochastic frame \textbf{P}rediction (RSP), a framework that learns visual representations from videos via stochastic future frame prediction.
We first describe how we revisit the idea of stochastic video generation \citep{denton2018stochastic} for representation learning and improve it by incorporating a recent recipe for self-supervised learning (see \cref{sec:representation_learning_from_videos}).
We then describe how we design a shared decoder architecture to effectively incorporate an auxiliary masked autoencoding objective \citep{he2022masked} that learns dense information within the static parts of each frame (see \cref{sec:representation_learning_from_images}). We provide the overview and pseudo-code of our framework in \cref{fig:method_overview} and \cref{algorithm:method}, respectively.

\subsection{Representation Learning from Videos with Stochastic Frame Prediction}
\label{sec:representation_learning_from_videos}
Our key idea is that learning a model that can predict multiple possible future frames can induce representations that capture temporal information between frames.
To this end, we build our framework upon the stochastic video generation \citep[SVG;][]{denton2018stochastic} model that captures uncertainty in future prediction by learning a time-dependent prior distribution over future frames.
Our key contribution lies in re-configuring SVG for representation learning by exploring multiple design choices and adopting recent advances in architectures and training techniques \citep{dosovitskiy2020image,hafner2020mastering,he2022masked,gupta2023siamese}, which we describe in the rest of this section.

\paragraph{Inputs and encoder} 
Given a video $\mathbf{x}$, we randomly sample two frames $\{\mathbf{x}_{t}, \mathbf{x}_{t+k}\}$ where $k$ is randomly chosen from a fixed set of values by following \citet{gupta2023siamese}.
Then we use the same vision transformer \citep[ViT;][]{dosovitskiy2020image} encoder $f^{\tt{enc}}_{\theta}$ that shares parameters for encoding frames $\mathbf{x}_{t}$ and $\mathbf{x}_{t+k}$.
Specifically, we extract non-overlapping patches from a frame, add 2D fixed sin-cos positional embeddings \citep{chen2021empirical}, and concatenate a \texttt{[CLS]} token to patches.
We note that we separately process each frame and do not concatenate patches from both frames.
We then process them through a series of Transformer layers \citep{vaswani2017attention} to obtain to obtain $\mathbf{h}_{t}$ and $\mathbf{h}_{t+k}$ consisting of \texttt{[CLS]} and patch representations.
\begin{gather}
\begin{aligned}
&\text{Encoder:} && &&\begin{aligned}\raisebox{2.05ex}{\llap{\blap{\ensuremath{ \hspace{0.1ex} \begin{cases} \hphantom{T} \\ \hphantom{T} \end{cases} \hspace*{-4ex}
}}}}
&\mathbf{h}_{t+k} = f^{\tt{enc}}_{\theta}(\mathbf{x}_{t+k})\\
&\mathbf{h}_{t} = f^{\tt{enc}}_{\theta}(\mathbf{x}_{t})
\end{aligned}
\label{eq:encoder}
\end{aligned}
\end{gather}

\makeatletter
\lst@Key{spacestyle}{}
  {\def\lst@visiblespace{{#1\lst@ttfamily{\char32}\textvisiblespace{}}}}
\makeatother

\lstset{
  backgroundcolor=\color{white},
  basicstyle=\fontsize{7.5pt}{7.5pt}\ttfamily\selectfont,
  columns=fullflexible,
  breaklines=true,
  captionpos=b,
  commentstyle=\fontsize{7.5pt}{7.5pt}\color{codeblue},
  keywordstyle=\fontsize{7.5pt}{7.5pt}\color{codekw},
  showspaces=true,
  showstringspaces=false,
  spacestyle   = \color{white},
}

\begin{algorithm}[t]
\caption{RSP: PyTorch-like Pseudocode}\label{algorithm:method}
\label{alg:code}
\definecolor{codeblue}{rgb}{0.25,0.5,0.5}
\definecolor{codekw}{rgb}{0.85, 0.18, 0.50}
\begin{lstlisting}[language=python]
# f, g: encoder, decoder
# q, p: posterior, learned prior

# input: x1 (current frame), x2 (future frame)
def rsp(x1, x2):
    h1, h2 = f(x1), f(perturb(x2))

    # Posterior distribution from both frames
    post_logits = q(cat(h1[:,0], h2[:,0]))
    post_dist   = make_dist(post_logits)
    post_z      = post_dist.rsample()

    # Prior distribution only from the current frame
    prior_logits = p(h1[:,0])
    prior_dist   = make_dist(prior_logits)

    pred_fut  = g(q=<mask>, kv=cat(h1, post_z))
    pred_loss = ((pred_fut - x2) ** 2).mean()
    kl_loss   = kl(post_dist, prior_dist)

    # Auxiliary MAE objective
    hm, mask, ids_restore = f(x2, mask=0.75)
    pred_mask = g(q=<mask>,        
                  kv=restore(hm, ids_restore))
    mae_loss  = ((pred_mask - x2) ** 2).mean(dim=-1)
    mae_loss  = (mae_loss * mask).sum() / mask.sum()
    
    loss = pred_loss + kl_scale * kl_loss + mae_loss
    return loss
    
\end{lstlisting}
\end{algorithm}

\paragraph{Augmentations}
We apply the same augmentation \textit{i.e.,} random resized crop and random horizontal flip, to both frames $\mathbf{x}_{t}$ and $\mathbf{x}_{t+k}$.
This is because applying such a strong augmentation differently to frames can sometimes make the two frames be significantly different from each other (see \cref{table:same_augmentation} for supporting experiments).
We then add a small Gaussian noise $\varepsilon \sim \mathcal{N}(0, \sigma)$ to the future frame $\mathbf{x}_{t+k}$ to discourage the model from finding a shortcut that simply copies pixels from $\mathbf{x}_{t+k}$ for predicting $\hat{\mathbf{x}}_{t+k}$.

\paragraph{Posterior and learned prior}
Following \citet{denton2018stochastic}, our framework consists of two main components: (i) a \textit{future frame prediction model} that predicts $\hat{\mathbf{x}}_{t+k}$ conditioned on $\mathbf{h}_{t}$ and a latent variable $\mathbf{z}_{t+k}$, which captures the uncertainty over future, from a posterior distribution $q_\theta(\mathbf{z}_{t+k} \,|\,\mathbf{h}_{t},\mathbf{h}_{t+k})$ and (ii) a \textit{prior network} that learns to approximate $p_{\theta}(\mathbf{z}_{t+k} \,|\, \mathbf{h}_{t})$ without access to the future frame.
\begin{gather}
\begin{aligned}
&\text{Posterior:} &&\mathbf{z}_{t+k}\sim q_\theta(\mathbf{z}_{t+k} \,|\,\mathbf{h}_{t},\mathbf{h}_{t+k}) \\
&\text{Learned prior:} &&\hat{\mathbf{z}}_{t+k}\sim p_\theta(\hat{\mathbf{z}}_{t+k} \,|\,\mathbf{h}_{t})
\label{eq:posterior_and_learned_prior}
\end{aligned}
\end{gather}
In our implementation, we introduce two small 2-layer MLP models that take \texttt{[CLS]} representations from both $\mathbf{h}_{t}$ and $\mathbf{h}_{t+k}$ for the posterior network and \texttt{[CLS]} representation from $\mathbf{h}_{t}$ for the prior network.
For the latent variable $\mathbf{z}_{t+k}$, we use a set of categorical variables by following \citet{hafner2020mastering} and use the straight-through estimator \citep{bengio2013estimating} for updating the parameters, which we find to be more effective than using Gaussian distribution (see \cref{table:ablations1} for supporting experiments).

\paragraph{Decoder}
For decoding, we first project $\mathbf{h}_{t}$ and $\mathbf{z}_{t+k}$ with a linear layer and concatenate them to $[\mathbf{h}_{t}, \mathbf{z}_{t+k}]$. 
Our decoder block consists of a (i) cross-attention layer where \texttt{[MASK]} tokens attend to tokens from $[\mathbf{h}_{t}, \mathbf{z}_{t+k}]$ and (ii) self-attention layer where \texttt{[MASK]} tokens attend to each other.
After processing the inputs through a series of decoder blocks, the final projection layer maps the token representations into normalized pixel patches $\hat{\mathbf{x}}_{t+k}$ \citep{he2022masked}.
\begin{gather}
\begin{aligned}
&\text{Decoder:} &&\hat{\mathbf{x}}_{t+k}\sim p_\theta(\hat{\mathbf{x}}_{t+k} \,|\,\mathbf{h}_{t}, \mathbf{z}_{t+k})
\label{eq:decoder}
\end{aligned}
\end{gather}
Here, we note that our architecture resembles the cross-self decoder \citep{gupta2023siamese} where unmasked patches from $\mathbf{x}_{t+k}$ attend to $\mathbf{x}_{t}$ via cross-attention layers.
But our design differs in that there is no interaction between $\mathbf{x}_{t}$ and $\mathbf{x}_{t+k}$ in our cross-attention layer.
We adopt this design to be able to share the decoder parameters for multiple objectives by making \texttt{[MASK]} tokens attend to different types of inputs via cross-attention layers,
which allows for effectively incorporating both frame prediction and MAE objectives into our framework, which we describe in \cref{sec:representation_learning_from_images}.

\paragraph{Objective}
We train the future frame prediction model to provide accurate prediction $\hat{\mathbf{x}}_{t+k}$ while minimizing the KL divergence between the prior distribution $p_{\theta}(\mathbf{z}_{t+k} \,|\, \mathbf{x}_{t})$ and the posterior distribution $q_{\theta}(\mathbf{z}_{t+k} \,|\, \mathbf{x}_{t}, \mathbf{x}_{t+k})$ as below:
\begin{gather}
\begin{aligned}
\mathcal{L}(\theta) = \mathbb{E}_{q_{\theta}(\mathbf{z}_{t+k} | \mathbf{x}_{t}, \mathbf{x}_{t+k})}\Big[-\ln p_\theta(\mathbf{x}_{t+k} | \mathbf{x}_{t}, \mathbf{z}_{t+k}&)\\
    +\beta\,\text{KL}\big[ q_{\theta}(\mathbf{z}_{t+k} | \mathbf{x}_{t}, \mathbf{x}_{t+k}) \,\Vert\,  p_{\theta}(\mathbf{z}_{t+k} | \mathbf{x}_{t}&) \big] 
    \Big],
\label{eq:objective}
\end{aligned}
\end{gather}
where $\beta$ is a loss scale hyperparameter that adjusts the balance between decoding loss and KL loss.
Intuitively, making the prior distribution to be closer to the posterior distribution corresponds to learning the prior network to \textit{predict the future}.
On the other hand, enabling the prediction model to generate better frames while making the posterior distribution closer to the prior distribution corresponds to \textit{making the latent variable more predictable} by the prior network \citep{denton2018stochastic}.
We find that our objective allows for learning strong representations from complex real-world videos when compared to the deterministic frame prediction model (see \cref{table:deterministic_prediction} for supporting experiments).

\subsection{Auxiliary Representation Learning from Images}
\label{sec:representation_learning_from_images}
While stochastic future frame prediction can induce representations capturing temporal information, it might focus less on the static parts of frames as the model has full access to the previous frame $\mathbf{x}_{t}$ when predicting $\mathbf{x}_{t+k}$.
To mitigate this issue, we introduce an auxiliary masked autoencoding \citep[MAE;][]{he2022masked} objective that focuses on learning the dense information within each frame.
Moreover, we design our framework to share the decoder across the frame prediction and MAE objectives, which enables both objectives to be synergistic with a small computational overhead.

\begin{figure} [t!]
    \centering
    \includegraphics[width=0.14\textwidth]{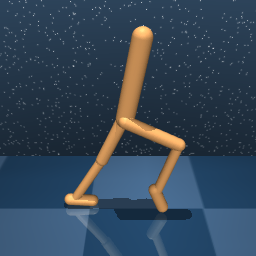}
    \includegraphics[width=0.14\textwidth]{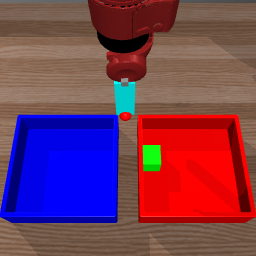}
    \includegraphics[width=0.14\textwidth]{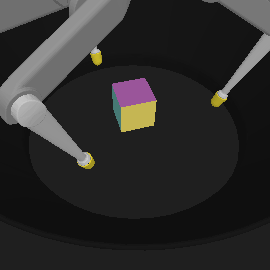}
    \\
    \includegraphics[width=0.14\textwidth]{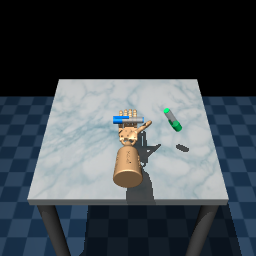}
    \includegraphics[width=0.14\textwidth]{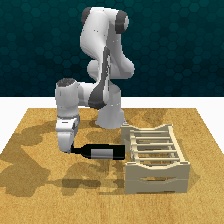}
    \includegraphics[width=0.14\textwidth]{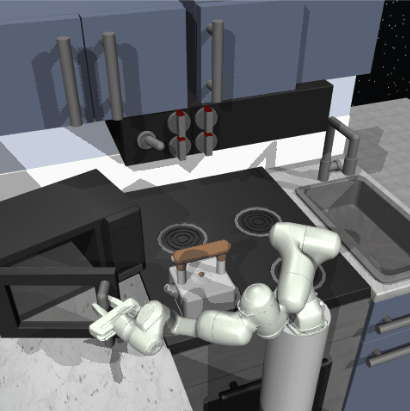}
    
    \vspace{-0.07in}
    \caption{
    Examples of visual observations from CortexBench \citep{majumdar2023we}, RLBench \citep{james2020rlbench}, and FrankaKitchen \citep{gupta2019relay}, which we used for training imitation learning agents that learn a mapping from observations to expert actions.
    Learning such agents requires representations that can understand both temporal and dense information.
    }
    \label{fig:task_example}
    \vspace{-0.1in}
\end{figure}
\paragraph{Masked autoencoding with shared decoder} We mask $m\%$ of the patches from $\mathbf{x}_{t+k}$ and process them through the encoder $f^{\tt{enc}}_{\theta}$ to obtain $\mathbf{h}_{t+k}^{m}$ consisting of \texttt{[CLS]} and unmasked patch representations.
We then project $\mathbf{h}_{t+k}^{m}$ with a linear layer, which is different from the linear layer used in the frame prediction, and process them through the \textit{shared} decoder by making \texttt{[MASK]} tokens attend to $\mathbf{h}_{t+k}^{m}$ via cross-attention layers.
Then the final projection layer maps the outputs into normalized pixel patches $\hat{\mathbf{x}}_{t+k}$.
\begin{gather}
\begin{aligned}
&\text{Masking:} &&\mathbf{x}^{m}_{t+k} \sim p^{\tt{mask}}(\mathbf{x}_{t+k}, m)\\
&\text{Encoder:} &&\mathbf{h}^{m}_{t+k} = f^{\tt{enc}}_{\theta}(\mathbf{x}^{m}_{t+k})\\
&\text{Decoder:} &&\hat{\mathbf{x}}_{t+k}\sim p_\theta(\hat{\mathbf{x}}_{t+k} \,|\,\mathbf{h}^{m}_{t+k})
\label{eq:svg}
\end{aligned}
  \end{gather}
We note that this auxiliary objective effectively enhances performance by complementing the frame prediction objective, with a negligible increase in training time.
We also empirically find that our shared decoder is crucial in making two objectives synergistic; training with a parallel decoder design achieves worse performance (see \cref{table:effect_auxiliary_mae} for supporting experimental results).

\begin{table*}[t]
\centering
\caption{\textbf{Results on vision-based robot learning.} Performance of imitation learning agents on CortexBench \citep{majumdar2023we} and RLBench \citep{james2020rlbench}, which are trained upon representations from ViT-S/16 model pre-trained on Kinetics-400 \citep{kay2017kinetics} dataset. We report the normalized score for DMC and success rates (\%) for other tasks.}
\vspace{-0.1in}
\label{table:main_robot_learning}
\begin{adjustbox}{max width=\textwidth}
\begin{tabular}{lcccccccccc}
\toprule
& \multicolumn{4}{c}{CortexBench} & \multicolumn{6}{c}{RLBench} \\
\cmidrule(lr){2-5} \cmidrule(lr){6-11}
Method & Adroit & MetaWorld & DMC & Trifinger & Button & Saucepan & Phone & Umbrella & Wine & Rubbish \\
\midrule
SimCLR \cite{chen2020simple} & 40.4 & 78.4 & 39.7 & 63.3 & \phantom{0}7.4 & 39.5 & 34.6 & \phantom{0}5.8 & 11.0 & \phantom{0}5.2 \\
MoCo v3 \cite{chen2021empirical} & 39.6 & 65.4 & 43.7 & 53.3 & 11.4 & 45.8 & 36.2 & 13.2 & \phantom{0}8.7 & \phantom{0}6.7 \\
Dino \cite{caron2021emerging}     & \bf{45.6} & 82.4 & 50.9 & 64.2 & 24.7 & 57.9 & 32.0 & 28.1 & 31.4 & 12.9 \\
MAE \cite{he2022masked}  & 44.8 & 81.4 & 52.1 & 62.2 & \phantom{0}6.4 & 36.8 & 37.7 & 10.0 & 10.0 & \phantom{0}6.2 \\
SiamMAE \cite{gupta2023siamese} & 44.0 & 81.1 & 56.0 & 52.1 & \phantom{0}6.1 & 22.5 & \phantom{0}5.4 & \phantom{0}4.0 & \phantom{0}8.7 & \phantom{0}3.5 \\
\textbf{RSP (Ours)}  & \bf{45.6} & \bf{84.5} & \bf{61.6} & \bf{66.2} & \bf{28.4} & \bf{93.4} & \bf{48.0} & \bf{37.3} & \bf{31.9} & \bf{18.5} \\
\bottomrule
\end{tabular}
\end{adjustbox}
\end{table*}
\section{Experiments}
In this section, we demonstrate the effectiveness of the proposed framework through evaluations on a variety of vision-based robot learning tasks including robotic manipulation and locomotion (see \cref{sec:robot_learning}) and video label propagation tasks including video segmentation and pose tracking (see \cref{sec:video_label_propagation}).
We also provide extensive ablation studies and analysis on our design choices (see \cref{sec:ablation_study_and_analysis}).

\subsection{Experimental Setup}

\paragraph{Pre-training}
For a fair comparison, we report all the experimental results using the ViT-S/16 model pre-trained on Kinetics-400 datasets \citep{kay2017kinetics} for 400 epochs.
We use the repeated sampling of 2 and count the epochs as effective epochs \citep{hoffer2020augment,feichtenhofer2022masked}.
For sampling frames $\mathbf{x}_{t}$ and $\mathbf{x}_{t+k}$, we follow  \citet{gupta2023siamese} that randomly samples $k$ from 4 to 48.
We implement our decoder block to sequentially have self-attention, cross-attention, and feedforward layers.
For the MAE objective, we use a 75\% masking ratio \citep{he2022masked}.
We use AdamW optimizer \citep{loshchilov2018decoupled} with a batch size of 1536.
For all baselines, we use the default hyperparameters.
We provide more details in \cref{appendix:implementation_details}.

\paragraph{Baselines}
We first consider image representation learning approaches, \textit{i.e.,} SimCLR \citep{chen2020simple}, MoCo v3 \citep{chen2021empirical}, Dino \citep{caron2021emerging} and MAE \citep{he2022masked}, as our baselines to compare our framework against standard image representation learning methods.
Moreover, we consider SiamMAE \citep{gupta2023siamese} as our baseline for its superior performance over other masked visual modeling methods \citep{feichtenhofer2022masked,tong2022videomae} and its resemblance to our approach.
With this comparison against SiamMAE, we evaluate the benefit of our stochastic frame prediction framework compared to the idea of predicting the masked patches of future frames conditioned on the unmasked patches.

\subsection{Vision-Based Robot Learning}
\label{sec:robot_learning}
We evaluate our framework on vision-based robot learning benchmarks, where the goal is to train imitation learning agents that solve target tasks by learning the mapping from visual observations to expert actions via behavior cloning \citep{pomerleau1988alvinn}.
We consider this setup because training such agents requires representations that capture both temporal and dense information from the visual observations (see \cref{fig:task_example} for examples of tasks used in our experiments).

\begin{figure}[t]
\centering
\vspace{-0.1in}
\includegraphics[width=0.48\textwidth]{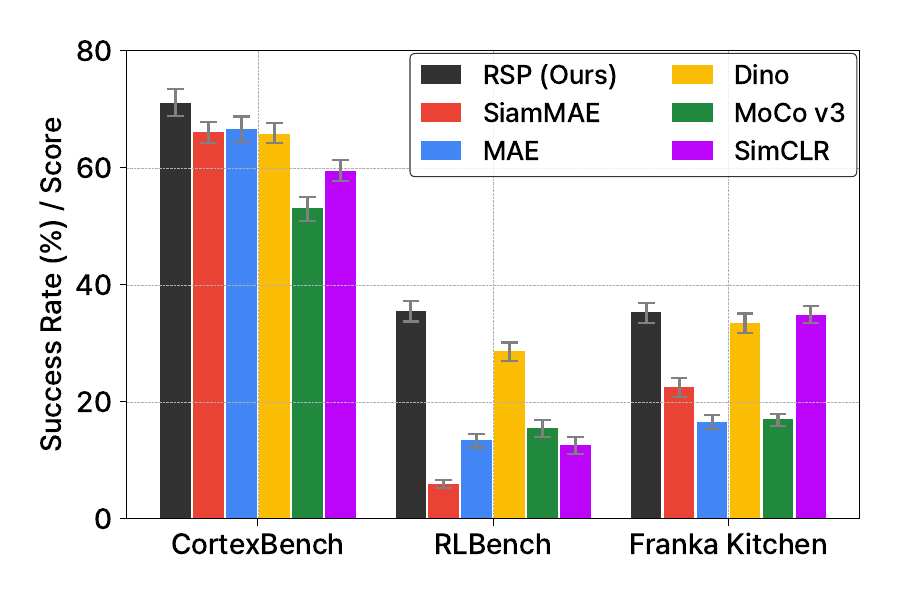}
\vspace{-0.375in}
\caption{\textbf{Aggregate results on vision-based robot learning.} We report the interquartile mean \citep{agarwal2021deep} over 20 vision-based robot learning tasks from CortexBench \citep{majumdar2023we}, RLBench \citep{james2020rlbench}, and Frana Kitchen \citep{gupta2019relay}.} 
\label{fig:main_robot_learning}
\end{figure}
\begin{table*}[t]
\centering
\caption{\textbf{Results on video label propagation.} We report performances on video segmentation, video part segmentation, and pose tracking tasks from DAVIS \citep{pont20172017}, VIP \citep{zhou2018adaptive}, and JHMDB \citep{jhuang2013towards} benchmarks, respectively.
For all methods, we report the performance with the representations pre-trained on the Kinetics-400 \citep{kay2017kinetics} dataset for 400 epochs. We further provide the performance of representations pre-trained on the ImageNet \citep{deng2009imagenet} dataset as a reference in \cref{appendix:results_with_imagenet_as_a_reference}.}
\label{table:main_video_label_propagation}
\vspace{-0.05in}
\begin{adjustbox}{max width=\textwidth}
\begin{tabular}{lccccccc}
\toprule
& & \multicolumn{3}{c}{DAVIS} & VIP & \multicolumn{2}{c}{JHMDB} \\
\cmidrule(lr){3-5} \cmidrule(lr){6-6} \cmidrule(lr){7-8}
Method & Architecture & $\gJ \& \gF_m$ & $\gJ_m$ & $\gF_m$ & mIoU & PCK@0.1 & PCK@0.2 \\
\midrule
SimCLR \cite{chen2020simple} & ViT-S/16 & 53.9 & 51.7 & 56.2 & 31.9 & 37.9 & 66.1 \\
MoCo v3 \cite{chen2021empirical} &  ViT-S/16 &  57.7 & 54.6 & 60.8 & 32.4 & 38.4 & 67.6 \\
Dino \cite{caron2021emerging}     &  ViT-S/16 & 59.5 & 56.5 & 62.5 & 33.4 & 41.1 & 70.3 \\
MAE \cite{he2022masked} &  ViT-S/16 & 53.5 & 50.4 & 56.7 & 32.5 & 43.0 & 71.3 \\
SiamMAE \cite{gupta2023siamese} &  ViT-S/16 & 58.1 & 56.6 & 59.6 & 33.3 & \textbf{44.7} & 73.0 \\
\textbf{RSP (Ours)}  &  ViT-S/16 & \textbf{60.1} & \textbf{57.4} & \textbf{62.8} & \textbf{33.8} & 44.6 & \textbf{73.4} \\
\midrule
\textbf{RSP (Ours)} & ViT-B/16 & 60.5 & 57.8 & 63.2 & 34.0 & 46.0 & 74.6 \\
\bottomrule
\end{tabular}
\end{adjustbox}
\vspace{0.08in}
\end{table*}

\paragraph{Experimental setup}
We first consider 4 domains from CortexBench \citep{majumdar2023we} which includes locomotion and manipulation tasks from various benchmarks \citep{rajeswaran2017learning,yu2020meta,tassa2020dm_control,bauer2022real}.
Moreover, we consider a more challenging setup by evaluating our framework on 6 manipulation tasks from RLBench \citep{james2020rlbench} which has successfully served as a simulation for sim-to-real transfer \citep{seo2023multi} or a proxy for real-robot experiments \citep{james2022coarse,shridhar2023perceiver}.
We train the imitation learning agents using 100 demos for each task, use keypoint augmentation \citep{james2022q} for demonstrations, and use the end-effector controller with path planning as an action mode. We use the front camera of 224$\times$224 resolution without depth for the CortexBench and RLBench.
Furthermore, we evaluate RSP on 5 tasks from Franka Kitchen \citep{gupta2019relay}, following the setup in \citet{nair2022r3m} that uses a left or right camera of 224$\times$224 resolution without depth.
For all the tasks, we follow the setup in \citet{majumdar2023we} that trains the agents upon \texttt{[CLS]} representation to predict expert actions.
We evaluate the model multiple times throughout training with a pre-defined interval and report the best performance.

\paragraph{Results}
We provide the main experimental results for each individual task (see \cref{table:main_robot_learning}) and aggregate performance (see \cref{fig:main_robot_learning}).
We first find that our framework outperforms all the baselines by a significant margin, as shown in \cref{fig:main_robot_learning} that reports interquartile mean \citep{agarwal2021deep} computed over 25 tasks from the benchmarks.
This demonstrates that our framework indeed can induce representations that could be useful for solving complex robot learning tasks that require temporal understanding.
We also observe that overall success rates are low in RLBench, as we consider a difficult setup of using only a single camera without depth information.
Nevertheless, we find our method consistently achieves superior performance to all the baselines.
In particular, RSP outperforms SiamMAE by a large margin in both benchmarks, \textit{i.e.,} RSP achieves 35.6\% while SiamMAE achieves  6.0\% success rates in RLBench.
This highlights the benefit of our approach that captures uncertainty over the future for representation learning.

\begin{figure*}[t!]
\centering
\includegraphics[width=0.99\textwidth]{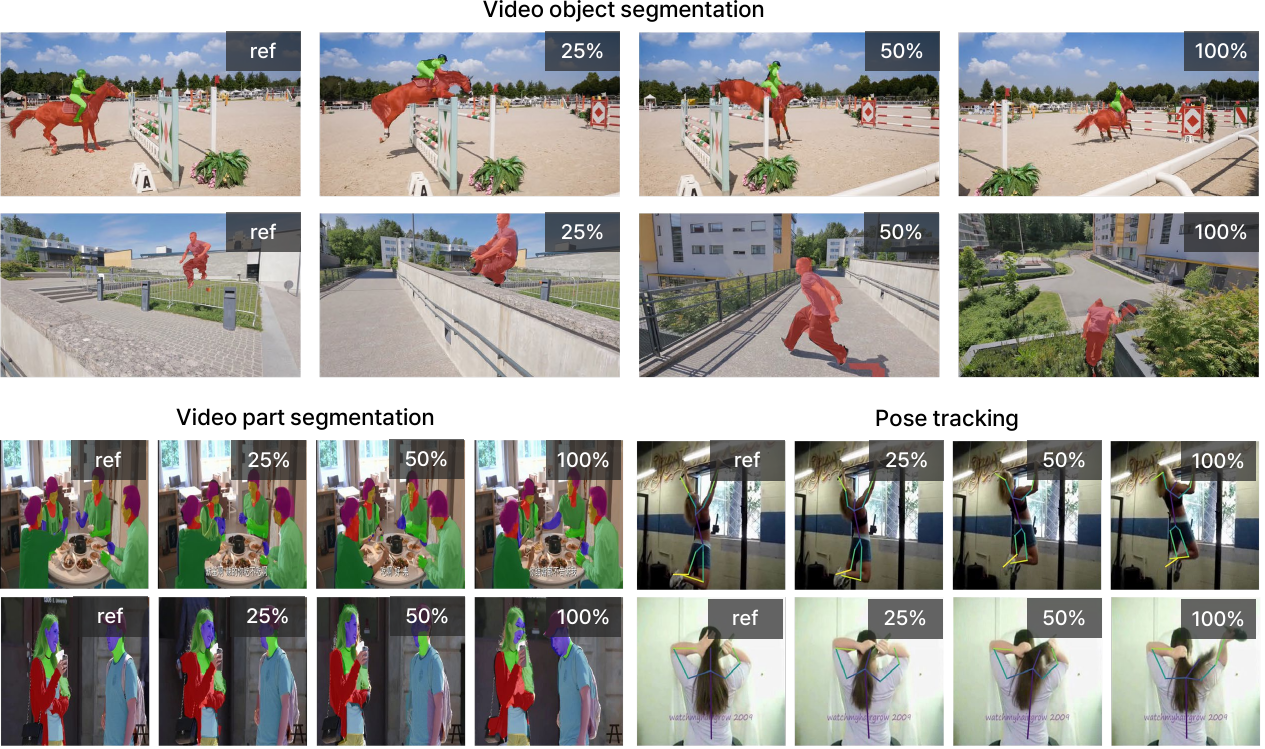}
\vspace{-0.05in}
\caption{\textbf{Qualitative results.} 
We provide examples of predicted propagation from RSP on video object segmentation \citep{pont20172017}, video part segmentation \citep{zhou2018adaptive}, and pose tracking \citep{jhuang2013towards} benchmarks. "ref" indicates the ground-truth annotations, and 25, 50, and 100\% refers to the propagated ratio of the videos. We provide additional qualitative results in \cref{appendix:additional_qualitative_results}.}
\label{fig:qualitative}
\vspace{-0.05in}
\end{figure*}

\subsection{Video Label Propagation}
\label{sec:video_label_propagation}
To evaluate how learned representations can capture temporal information between frames, we report the performance of three video label propagation tasks.
The goal of these tasks is, given a first frame with ground-truth annotations, to predict the labels in each pixel from future frames.

\begin{table*}[t]
\centering
\subfloat[
\textbf{Deterministic prediction.} We find our stochastic future frame prediction objective is crucial for representation learning from complex real-world videos.
\label{table:deterministic_prediction}
]{
\centering
\begin{minipage}{0.45\linewidth}
\begin{center}
\scriptsize
\tablestyle{3pt}{1.2}
\begin{tabular}{cccc}
Stochastic & $\mathcal{J}$\&$\mathcal{F}_m$ & $\mathcal{J}_m$ & $\mathcal{F}_m$ \\ 
\shline
\xmark & 54.4 & 50.7 & 58.1 \\
\baseline{\cmark} & \baseline{\textbf{60.1}} & \baseline{\textbf{57.4}} & \baseline{\textbf{62.8}} \\
\end{tabular}
\end{center}
\end{minipage}
}
\hspace{1em}
\subfloat[
\textbf{Stochastic latent variable.} Using a set of categorical variables \citep{hafner2020mastering} for the latent variable outperforms a baseline that employs Gaussian distribution.
\label{table:stochastic_latent_variable}
]{
\centering
\begin{minipage}{0.45\linewidth}
\begin{center}
\scriptsize
\tablestyle{3pt}{1.2}
\begin{tabular}{cccc}
Latent & $\mathcal{J}$\&$\mathcal{F}_m$ & $\mathcal{J}_m$ & $\mathcal{F}_m$ \\ 
\shline
Gaussian & 54.1 & 52.9 & 55.9 \\
\baseline{Categorical} & \baseline{\textbf{60.1}} & \baseline{\textbf{57.4}} & \baseline{\textbf{62.8}} \\
\end{tabular}
\end{center}
\end{minipage}
}
\\[2mm]
\subfloat[
\textbf{Auxiliary MAE objective with a shared decoder.} 
We find that training with the auxiliary MAE objective works better, especially when combined with our shared decoder design.
\label{table:effect_auxiliary_mae}
]{
\centering
\begin{minipage}{0.45\linewidth}
\begin{center}
\scriptsize
\tablestyle{3pt}{1.2}
\begin{tabular}{ccccc}
w/ MAE & Decoder & $\mathcal{J}$\&$\mathcal{F}_m$ & $\mathcal{J}_m$ & $\mathcal{F}_m$ \\ 
\shline
\xmark &  - & 57.7 & 54.9 & 60.5 \\
\cmark &  Separate & 58.1 & 55.4 & 60.7 \\
\baseline{\cmark} &  \baseline{Shared} &\baseline{\textbf{60.1}} & \baseline{\textbf{57.4}} & \baseline{\textbf{62.8}}  \\
\end{tabular}
\end{center}
\end{minipage}
}
\hspace{1em}
\subfloat[
\textbf{KL objective scale.} Training a model with too strong or weak KL objectives leads to worse performance.
\label{table:kl_loss_scale}
]{
\centering
\begin{minipage}{0.45\linewidth}
\begin{center}
\scriptsize
\tablestyle{5pt}{1.2}
\begin{tabular}{cccc}
KL scale & $\mathcal{J}$\&$\mathcal{F}_m$ & $\mathcal{J}_m$ & $\mathcal{F}_m$ \\ 
\shline
0.1 & 56.1 & 52.9 & 59.3 \\
\baseline{\textbf{0.01}} & \baseline{\textbf{60.1}} & \baseline{\textbf{57.4}} & \baseline{\textbf{62.8}}  \\
0.001 & 59.1 & 56.6 & 61.5 \\
\end{tabular}
\end{center}
\end{minipage}
}
\vspace{-0.025in}
\caption{\textbf{Ablation studies.} We report the performance of various variants of RSP on DAVIS benchmark. For all experiments, we pre-train ViT-S/16 model on Kinetics-400 dataset for 400 epochs. Default settings are highlighted in \colorbox{baselinecolor}{gray}.}
\label{table:ablations1} 
\vspace{-0.15in}
\end{table*}
\paragraph{Experimental setup}
We consider the video object segmentation, video part segmentation, and pose tracking tasks from DAVIS \citep{pont20172017}, VIP \citep{zhou2018adaptive}, and JHMDB \citep{jhuang2013towards} benchmarks, respectively.
For evaluation, we follow the protocol of prior work \citep{wang2019learning,li2019joint,lai2019self,jabri2020space} that uses a $k$-nearest neighbor inference, maintain a queue of length $m$ to provide a temporal context and use a restricted set of source nodes with a spatial radius $r$.
Due to computational constraints, we compare our framework against the baselines pre-trained under the same budget using the same architecture of ViT-S/16.
We conduct a grid search on evaluation hyperparameters for each method and report the best performance.

\paragraph{Results}
We provide the quantitative evaluation in \cref{table:main_video_label_propagation} and qualitative results in \cref{fig:qualitative}.
As shown in \cref{table:main_video_label_propagation},
we find that our framework achieves superior or competitive performance to all the baselines in video label propagation tasks.
In particular, our framework, with both stochastic frame prediction and auxiliary MAE objectives, outperforms MAE by a large margin, \textit{i.e.,} 6.6\%p.
This highlights the effectiveness of stochastic future frame prediction objectives for temporal understanding.
Moreover, similar to the trend from robot learning experiments in \cref{sec:robot_learning}, we find our framework outperforms SiamMAE.
This again demonstrates the benefit of our approach over masked visual modeling approaches for image representation learning from videos.

\subsection{Ablation Study and Analysis}
\label{sec:ablation_study_and_analysis}
We provide extensive ablation studies and analysis to investigate the importance of our design choices for building our framework upon prior work \citep{denton2018stochastic}.
Due to computational constraints, we report the performance on the DAVIS benchmark.

\paragraph{Comparison with deterministic frame prediction}
To investigate the importance of \textit{stochastic} future prediction, we compare our framework with \textit{deterministic} frame prediction model. For a fair comparison, we also use the auxiliary MAE objective with the shared decoder for both methods.
In \cref{table:deterministic_prediction}, we find that the deterministic frame prediction model significantly underperforms our framework, \textit{i.e.,} the deterministic baseline achieves 54.4\% while our stochastic framework achieves 60.1\%.
This shows that deterministic frame predictor struggles to learn useful representations from complex large video datasets like Kinetics-400 \citep{kay2017kinetics}.
On the other hand, our method can learn such representations by learning to predict possible multiple futures via stochastic frame prediction.

\paragraph{Latent variable design}
We explore two design choices on the stochastic latent variable $\mathbf{z}_{t+k}$. Specifically, we consider two variants that employ \textit{Gaussian} distribution or a set of \textit{Categorical} variables \citep{hafner2020mastering}.
Interestingly, in \cref{table:stochastic_latent_variable}, we find that utilizing the Categorical variable significantly outperforms the variant with Gaussian distribution.
We hypothesize this is because it is easier to predict discrete labels compared to accurately approximating continuous Gaussian distribution.
In addition, we would like to note that \citet{meyer2023harnessing} demonstrated RL with discrete representations outperforms continuous representations when the environment dynamics gets more complex. This could also explain our observation because the Kinetics-400 \citep{kay2017kinetics} dataset consists of complex real-world videos.
Given this result, it would be an interesting future direction to design models with a more expressive prior, \textit{e.g.,} autoregressive prior.

\begin{table*}[t]
\centering
\subfloat[
\textbf{Applying the same augmentation.} Applying augmentations (\textit{i.e.,} random resized crop and horizontal flip) differently to current and future frames significantly degrades performance.
\label{table:same_augmentation}
]{
\centering
\begin{minipage}{0.45\linewidth}
\begin{center}
\scriptsize
\tablestyle{3pt}{1.2}
\begin{tabular}{cccc}
Same aug & $\mathcal{J}$\&$\mathcal{F}_m$ & $\mathcal{J}_m$ & $\mathcal{F}_m$ \\ 
\shline
\xmark & 53.7 & 52.2 & 55.2 \\
\baseline{\cmark} & \baseline{\textbf{60.1}} & \baseline{\textbf{57.4}} & \baseline{\textbf{62.8}} \\
\end{tabular}
\end{center}
\end{minipage}
}
\hspace{1em}
\subfloat[
\textbf{Future frame augmentation.} Applying a mild augmentation to a future frame can enhance performance. But strong augmentation such as masking degrades the performance.
\label{table:future_frame_augmentation}
]{
\centering
\begin{minipage}{0.45\linewidth}
\begin{center}
\scriptsize
\tablestyle{3pt}{1.2}
\begin{tabular}{ccccc}
Future frame aug & Scale & $\mathcal{J}$\&$\mathcal{F}_m$ & $\mathcal{J}_m$ & $\mathcal{F}_m$ \\ 
\shline
None & - & 58.3 & 56.1 & 60.6 \\\midrule
Masking & 0.75 & 57.7 & 54.8 & 60.6\\
Masking & 0.95 & 55.8 & 52.7 & 58.9 \\\midrule
Noise & 0.1 & 58.4 & 56.0 & 60.7 \\
\baseline{Noise} & \baseline{0.5} & \baseline{\textbf{60.1}} & \baseline{\textbf{57.4}} & \baseline{\textbf{62.8}} \\
Noise & 1.0 & 58.9 & 56.3 & 61.4 \\
\end{tabular}
\end{center}
\end{minipage}
}
\vspace{-0.175in}
\caption{\textbf{Effect of data augmentation.} We investigate (a) the importance of applying the same augmentation to current and future frames and (b) the effect of applying mild augmentation to the future frame. Default settings are highlighted in \colorbox{baselinecolor}{gray}.}
\label{table:ablations2}
\vspace{-0.1in}
\end{table*}

\paragraph{Auxiliary MAE objective with shared decoder}
One important design in our framework is introducing the \textit{auxiliary MAE objective} to learn dense representation within the frames, which might not be learned by the frame prediction objective.
In \cref{table:effect_auxiliary_mae}, we observe that our framework indeed outperforms a baseline that does not introduce the auxiliary objective by a large margin (+2.4\%p).
Moreover, to investigate the importance of having a \textit{shared decoder}, we design a parallel decoder baseline that has an additional, separate decoder for the auxiliary MAE objective.
We find that having a shared decoder is crucial for making both objectives synergistic, \textit{i.e.,} our framework with the shared decoder achieves 60.1\% while the parallel decoder baseline achieves 58.1\%.
This result is intriguing because our shared decoder design also has the benefit of being parameter-efficient compared to the parallel decoder.

\begin{figure}[t]
\vspace{-0.2in}
\centering
\includegraphics[width=0.37\textwidth]{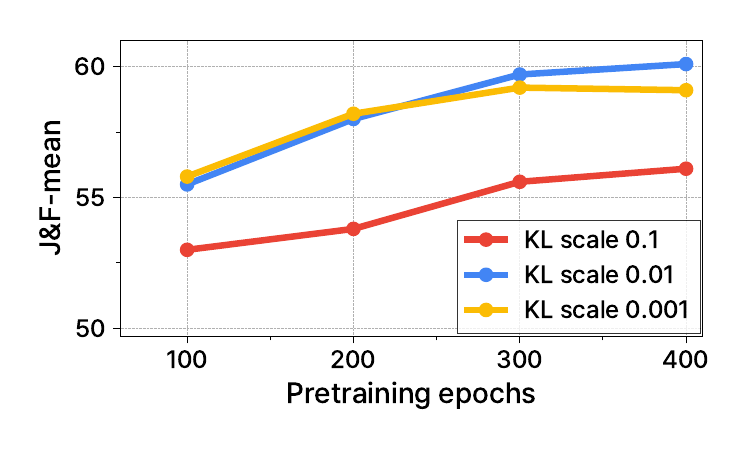}
\vspace{-0.22in}
\caption{\textbf{Effect of KL loss scale.} We report the learning curves of models trained with different KL loss scales ($\beta$).
}
\vspace{-0.2in}
\label{fig:main_kl_loss_scale}
\end{figure}

\paragraph{Effect of KL loss scale}
We also conduct analysis on the effect of the KL loss scale ($\beta$) to provide a deeper understanding of the learning dynamics of our framework.
In \cref{table:kl_loss_scale}, we observe that too strong or weak KL loss scales lead to worse performance.
This is because high $\beta$ makes it difficult to learn good posterior by enforcing distributions to be close too early, \textit{i.e.,} before the model starts to learn a good posterior distribution, which leads to overall worse performance as shown in \cref{fig:main_kl_loss_scale}.
On the other hand, low $\beta$ makes the posterior distribution tend to ignore the prior distribution, and this consequently makes it difficult for the prior model to predict the posterior, which leads to lower asymptotic performance as shown in \cref{fig:main_kl_loss_scale}.

\paragraph{Applying the same augmentation}
As we previously mentioned in \cref{sec:representation_learning_from_videos}, applying the same augmentation to both current and future frames is crucial for making the frame prediction objective valid.
For instance, applying the random horizontal flipping augmentation differently to current and future frames would make it impossible to predict the future frame.
In \cref{table:same_augmentation}, we indeed find that applying different augmentations to current and future frames significantly degrades the performance.

\paragraph{Additional future frame augmentation} 
We study the effect of our design choice that augments the future frame by adding a small Gaussian noise in \cref{table:future_frame_augmentation}.
We also explore another augmentation scheme of applying masks to future frames, similar to \citet{gupta2023siamese}.
We find that applying masking augmentation degrades the performance, exhibiting a similar trend in \cref{table:same_augmentation}.
This is because the prior have to also capture the stochasticity from aggressive masking augmentation, which makes it difficult to learn meaningful prior distribution.
On the other hand, adding a small Gaussian noise can effectively improve the performance by delivering the benefit of augmentation, as it does not change the semantic meaning of frames.

\section{Conclusion}
In this work, we present RSP, a framework for visual representation learning from videos, that learns representations that capture temporal information between frames by training a stochastic future frame prediction model.
Our key contribution lies in re-visiting the idea of stochastic video generation \citep{denton2018stochastic} and re-designing it for representation learning by exploring and adopting various design choices.
Our extensive experiments demonstrate that our framework consistently achieves competitive or superior performance to various baselines.
We hope our work further facilitates research on representation learning from videos via future frame prediction.

\paragraph{Limitations and future directions}
One limitation of our work is that the quality of generated frames is not of high quality, though our focus is not on high-fidelity generation.
Given this, it would be an interesting direction to incorporate recent video generative models based on diffusion models, similar to \citet{hudson2023soda} that learns representations via image diffusion models.
Moreover, due to computational constraints, our work does not include large-scale experiments with longer training budgets and larger models.
Scaling up our approach would be an interesting future direction.
Finally, an extension of our framework to multiple frames is a future direction we are keen to explore.

\section*{Impact Statement}
This paper presents a framework for representation learning via generative modeling of videos.
Thus there is a risk of potential misuse of our model for malicious purposes, \textit{e.g.,} generating fake videos.
However, unlike other high-fidelity generative models, our model generates outputs that are clearly distinguishable from real frames.
This significantly reduces the risk of our model being used for generating fake videos.
Nonetheless, it is still important to recognize and state such potential risk of misuse as the potential extension of our work is likely to have the capability to learn strong representations while generating high-quality videos.

\section*{Acknowledgements}
This work was supported by Institute of Information \& communications Technology Planning \& Evaluation (IITP) grant funded by the Korea government (MSIT) (No.RS-2019-II190075, Artificial Intelligence Graduate School Program (KAIST); No.RS-2021-II212068, Artificial Intelligence Innovation Hub and Samsung Electronics Co., Ltd (IO201211-08107-01). We also appreciate NVIDIA Corporation (\url{https://www.nvidia.com/}) for providing compute resources.

\newpage

\bibliography{main}
\bibliographystyle{icml2024}

\newpage
\appendix
\onecolumn

\section{Implementation Details}
\label{appendix:implementation_details}
We build our framework upon the official implementation of MAE \citep{he2022masked}.\footnote{\url{https://github.com/facebookresearch/mae}}
We summarize our hyperparameters of pre-training and video label propagation in \cref{tab:appendix_hparam}.
We follow \citet{hafner2021mastering} for various design choices with regard to stochastic latent variable.
Specifically, we employ a set of 32 Categorical variables with 32 classes for the posterior and prior distributions.
Furthermore, to prevent over-regularizing the representations towards an inadequately trained prior, we incorporate KL balancing with a ratio of $\alpha=0.8$, as introduced in \citet{hafner2021mastering}.

\begin{table}[h]
\centering
\subfloat[
\textbf{Pre-training hyperparameters}
]{
\centering
\begin{minipage}{0.6\linewidth}
\begin{center}
\scriptsize
\tablestyle{3pt}{1.2}
\begin{tabular}{lc}
config & value \\
\shline
optimizer & AdamW \citep{loshchilov2018decoupled} \\
optimizer momentum & $\beta_1$, $\beta_2$ = 0.9, 0.95 \citep{chen2020generative} \\
optimizer weight decay & 0.05 \\
learning rate & 1.5e-4 \\
learning rate scheduler & Cosine decay \citep{loshchilov2017sgdr}\\
warmup epochs \citep{goyal2017accurate} & 40 \\
pre-train epochs & 400 \\
repeated sampling \citep{hoffer2020augment} & 2 \\
batch size & 1536 \\
frame sampling gap & [4, 48] \\
augmentation & hflip, crop [0.5, 1.0] \\
Discrete latent dimensions & 32 \\
Discrete latent classes & 32 \\
KL balancing ratio & 0.8 \\
\end{tabular}
\end{center}
\end{minipage}
}
\hspace{1em}
\subfloat[
\textbf{Evaluation hyperparameters}
]{
\centering
\begin{minipage}{0.3\linewidth}
\begin{center}
\scriptsize
\tablestyle{3pt}{1.2}
\begin{tabular}{lccc}
config & DAVIS & VIP & JHMDB \\
\shline
top-k & 7 & 7 & 10 \\
neighborhood size & 30 & 5 & 5 \\
queue length & 30 & 3 & 30 \\
\end{tabular}
\end{center}
\end{minipage}
}
\caption{\textbf{Hyperparameter details of pre-training and evaluation}}
\label{tab:appendix_hparam}

\end{table}

\paragraph{Architectural details}
We use standard ViT-S/16 \citep{dosovitskiy2020image} as our encoder. For the decoder, each block is composed of cross-attention, self-attention, and feed-forward MLP layers. The hyperparameters for the decoder, including embedding dimension, depth, and the number of heads, are aligned with those specified in \citet{he2022masked}.

\section{Additional Ablation Study and Analysis}
\label{appendix:additional_ablation_study_and_analysis}
We provide additional ablation studies and analysis to investigate the importance of our design choices. We report the performance on the DAVIS benchmark in Table \ref{table:appendix_ablations}.
\begin{table*}[h]
\centering
\subfloat[
\textbf{Encoder-decoder projection.} We find that distinct projections for stochastic frame prediction and auxiliary MAE objective is crucial for learning representation.
\label{table:ablation_projection}
]{
\centering
\begin{minipage}{0.45\linewidth}
\begin{center}
\scriptsize
\tablestyle{3pt}{1.2}
\begin{tabular}{cccc}
Projection & $\mathcal{J}$\&$\mathcal{F}_m$ & $\mathcal{J}_m$ & $\mathcal{F}_m$ \\ 
\shline
Same & 56.6 & 54.3 & 58.9 \\
\baseline{Distinct} & \baseline{\textbf{60.1}} & \baseline{\textbf{57.4}} & \baseline{\textbf{62.8}} \\
\end{tabular}
\end{center}
\end{minipage}}
\hspace{1em}
\subfloat[
\textbf{Concatenating latent variable and patch representations.} We find that concatenating the latent variable and patch representations along the channel dimension works better than concatenating them along the channel dimension.
\label{table:ablation_concat}
]{
\centering
\begin{minipage}{0.45\linewidth}
\begin{center}
\scriptsize
\tablestyle{3pt}{1.2}
\begin{tabular}{cccc}
Concat & $\mathcal{J}$\&$\mathcal{F}_m$ & $\mathcal{J}_m$ & $\mathcal{F}_m$ \\ 
\shline
Channel dim & 54.1 & 52.9 & 55.9 \\
\baseline{Tokens} & \baseline{\textbf{60.1}} & \baseline{\textbf{57.4}} & \baseline{\textbf{62.8}} \\
\end{tabular}
\end{center}
\end{minipage}}
\vspace{-0.025in}
\caption{\textbf{Ablation studies.} We report the performance of various variants of RSP on DAVIS benchmark. For all experiments, we pre-train ViT-S/16 model on Kinetics-400 dataset for 400 epochs. Default settings are highlighted in \colorbox{baselinecolor}{gray}.}
\label{table:appendix_ablations} 
\vspace{-0.15in}
\end{table*}

\newpage

\section{Experimental Results with 95\% Confidence Interval}
\label{appendix:experimental_results_with_confidence_interval}
We here provide the experimental results of \cref{table:main_robot_learning} with 95\% confidence intervals in \cref{tab:appendix-95ci}.
\begin{table*}[h]
\centering
\caption{\textbf{Results on vision-based robot learning.} Performance of imitation learning agents on CortexBench \citep{majumdar2023we}, RLBench \citep{james2020rlbench}, and Franka Kitchen \citep{gupta2019relay} with a 95\% confidence interval. We have 5, 4, and 4 runs for CortexBench, RLBench, and Franka Kitchen respectively.}
\label{tab:appendix-95ci}
\subfloat[CortexBench]{
\begin{tabular}{lcccc}
\toprule
Method & Adroit & MetaWorld & DMC & Trifinger \\
\midrule
SimCLR & 40.4$\pm$3.3 & 78.4$\pm$5.2 & 39.7$\pm$2.9 & 63.3$\pm$3.3 \\
MoCo v3 & 39.6$\pm$4.3 & 65.4$\pm$8.0 & 43.7$\pm$3.2 & 53.3$\pm$1.6 \\
Dino     & \textbf{45.6}$\pm$6.2 & 82.4$\pm$5.8 & 50.9$\pm$1.5 & 64.2$\pm$3.5 \\
MAE & 44.8$\pm$4.3 & 81.4$\pm$6.3 & 52.1$\pm$3.7 & 62.2$\pm$5.0 \\
SiamMAE  & 44.0$\pm$6.6 & 81.1$\pm$6.3 & 56.0$\pm$2.9 & 52.1$\pm$7.6 \\
\textbf{RSP (Ours)}  & \textbf{45.6}$\pm$4.6 & \textbf{84.5}$\pm$6.6 & \textbf{61.6}$\pm$3.4 & \textbf{66.2}$\pm$0.8 \\
\bottomrule
\end{tabular}}
\vspace{0.1in}
\subfloat[RLBench]{
\begin{tabular}{lcccccccccc}
\toprule
Method & Button & Saucepan & Phone & Umbrella & Wine & Rubbish \\
\midrule
SimCLR & \phantom{0}7.4$\pm$2.6 & 39.5$\pm$2.2 & 34.6$\pm$6.6 & \phantom{0}5.8$\pm$3.3 & 11.0$\pm$2.1 & \phantom{0}5.2$\pm$1.2 \\
MoCo v3 & 11.4$\pm$4.1 & 45.8$\pm$3.9 & 36.2$\pm$3.4 & 13.2$\pm$1.5 & \phantom{0}8.7$\pm$0.7 & \phantom{0}6.7$\pm$0.8 \\
Dino     & 24.7$\pm$1.5 & 57.9$\pm$5.9 & 32.0$\pm$5.5 & 28.1$\pm$1.4 & 31.4$\pm$1.5 & 12.9$\pm$1.5 \\
MAE & \phantom{0}6.4$\pm$2.2 & 36.8$\pm$6.4 & 37.7$\pm$1.9 & 10.0$\pm$1.2 & 10.0$\pm$2.1 & \phantom{0}6.2$\pm$3.2 \\
SiamMAE  & \phantom{0}6.1$\pm$2.3 & 22.5$\pm$0.8 & \phantom{0}5.4$\pm$0.5 & \phantom{0}4.0$\pm$0.0 & \phantom{0}8.7$\pm$0.8 & \phantom{0}3.5$\pm$0.9 \\
\textbf{RSP (Ours)}  & \textbf{28.4}$\pm$3.0 & \textbf{93.4}$\pm$1.8 & \textbf{48.0}$\pm$4.6 & \textbf{37.3}$\pm$3.0 & \textbf{31.9}$\pm$2.3 & \textbf{18.5}$\pm$1.1 \\
\bottomrule
\end{tabular}
}
\vspace{0.1in}
\subfloat[Franka Kitchen]{
\begin{tabular}{lccccc}
\toprule
Method & Knob1 on & Light on & Sdoor open & Ldoor open & Micro open \\
\midrule
SimCLR  & 25.3$\pm$2.1 & \textbf{55.8}$\pm$6.4 & 72.3$\pm$2.8 & 17.0$\pm$2.9 & 23.3$\pm$2.8 \\
MoCo v3  & 11.5$\pm$3.9 & 24.3$\pm$5.0 & 66.5$\pm$3.2 & 10.3$\pm$2.1 & 14.3$\pm$2.5 \\
Dino  & 27.0$\pm$3.2 & 44.3$\pm$6.5 & 77.0$\pm$5.0 & 16.5$\pm$2.5 & 28.5$\pm$4.8 \\
MAE  & 12.0$\pm$3.3 & 24.3$\pm$4.2 & 71.5$\pm$4.3 & 12.8$\pm$3.9 & 10.0$\pm$2.8 \\
SiamMAE & 16.8$\pm$4.4 & 36.5$\pm$7.0 & 68.0$\pm$7.9 & 17.3$\pm$3.7 & 13.5$\pm$4.8 \\
\textbf{RSP (Ours)}  & \textbf{31.0}$\pm$2.4 & 44.5$\pm$5.6 & \textbf{82.5}$\pm$2.7 & \textbf{28.8}$\pm$4.8 & \textbf{30.3}$\pm$5.6 \\
\bottomrule
\end{tabular}}
\end{table*}

\newpage

\section{Comparison with ImageNet Pre-trained SSLs}
\label{appendix:results_with_imagenet_as_a_reference}

\begin{table*}[h]
\centering
\caption{\textbf{Results on video label propagation.} We report performances on video segmentation, video part segmentation, and pose tracking tasks from DAVIS \citep{pont20172017}, VIP \citep{zhou2018adaptive}, and JHMDB \citep{jhuang2013towards} benchmarks, respectively. We compare the Kinetics-400 pre-trained approaches to the ImageNet pre-trained approaches as a reference.
}
\label{table:appendix-main}
\vspace{-0.05in}
\begin{adjustbox}{max width=\textwidth}
\begin{tabular}{lccccccc}
\toprule
& & \multicolumn{3}{c}{DAVIS} & VIP & \multicolumn{2}{c}{JHMDB} \\
\cmidrule(lr){3-5} \cmidrule(lr){6-6} \cmidrule(lr){7-8}
Method & Architecture & $\gJ \& \gF_m$ & $\gJ_m$ & $\gF_m$ & mIoU & PCK@0.1 & PCK@0.2 \\
\midrule
\rowcolor{RowHighlight}
\multicolumn{8}{c}{\textit{Kinetics-400 pre-trained}} \\
\midrule
SimCLR \cite{chen2020simple} & ViT-S/16 & 53.9 & 51.7 & 56.2 & 31.9 & 37.9 & 66.1 \\
MoCo v3 \cite{chen2021empirical} &  ViT-S/16 &  57.7 & 54.6 & 60.8 & 32.4 & 38.4 & 67.6 \\
Dino \cite{caron2021emerging}     &  ViT-S/16 & 59.5 & 56.5 & 62.5 & 33.4 & 41.1 & 70.3 \\
MAE \cite{he2022masked} &  ViT-S/16 & 53.5 & 50.4 & 56.7 & 32.5 & 43.0 & 71.3 \\
SiamMAE \cite{gupta2023siamese} &  ViT-S/16 & 58.1 & 56.6 & 59.6 & 33.3 & \textbf{44.7} & 73.0 \\
\textbf{RSP (Ours)}  &  ViT-S/16 & \textbf{60.1} & \textbf{57.4} & \textbf{62.8} & \textbf{33.8} & 44.6 & \textbf{73.4} \\
\midrule
\textbf{RSP (Ours)} & ViT-B/16 & 60.5 & 57.8 & 63.2 & 34.0 & 46.0 & 74.6 \\
\midrule
\rowcolor{RowHighlight}
\multicolumn{8}{c}{\textit{ImageNet pre-trained}} \\
\midrule
Dino \cite{caron2021emerging} & ViT-S/16 & 61.8 & 60.2 & 63.4 & 36.2 & 45.6 & 75.0 \\
MAE \cite{he2022masked} & ViT-B/16 & 53.5 & 52.1 & 55.0 & 28.1 & 44.6 & 73.4 \\
\bottomrule
\end{tabular}
\end{adjustbox}
\vspace{0.08in}
\end{table*}

\newpage

\section{Additional Qualitative Results}
\label{appendix:additional_qualitative_results}

\begin{figure*}[h]
\centering
\includegraphics[width=0.99\textwidth]{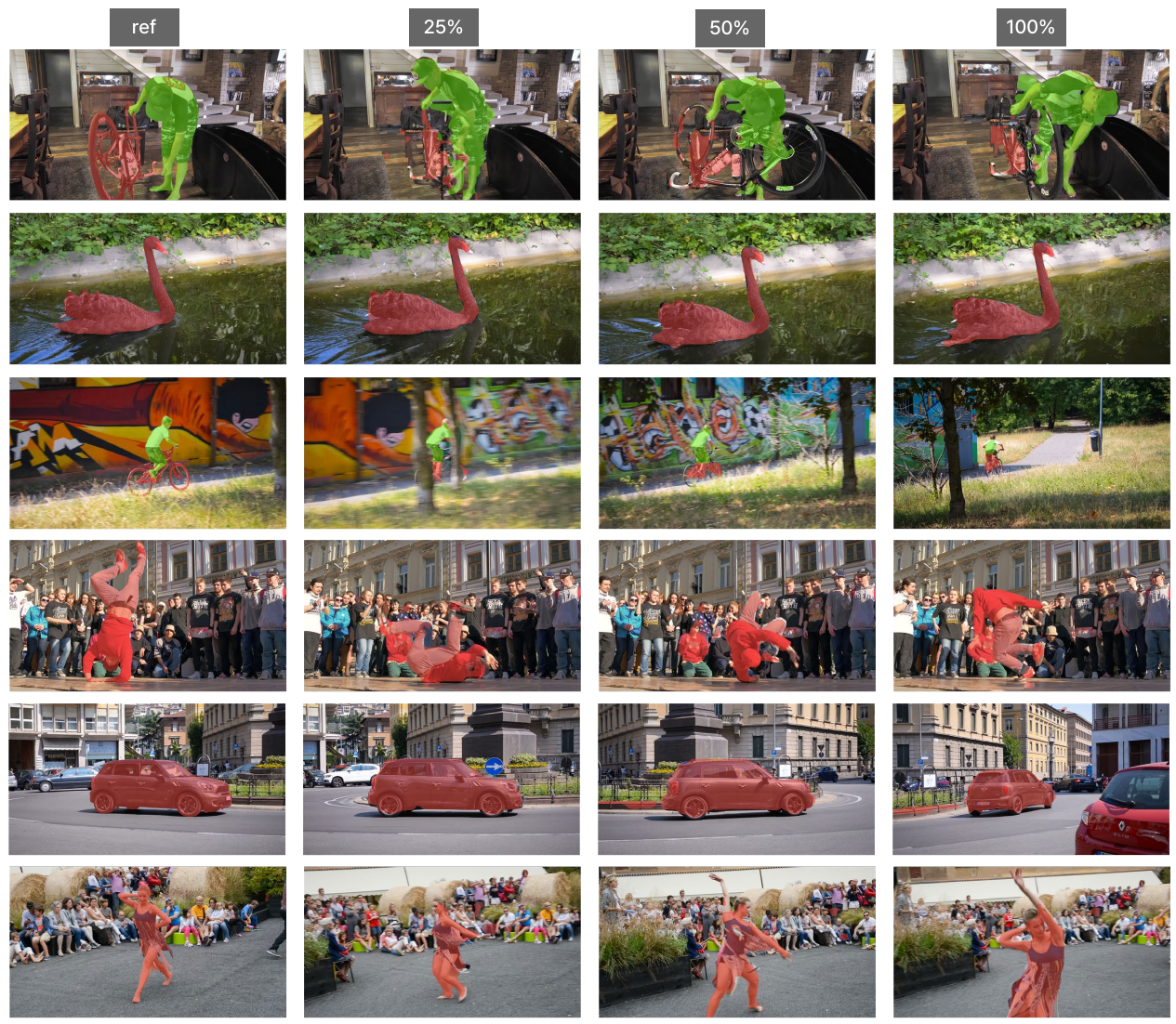}
\caption{\textbf{Additional qualitative results.} 
We provide more qualitative results of predicted propagation from RSP on DAVIS video object segmentation \citep{pont20172017} benchmarks. "ref" indicates the ground-truth annotations, and 25, 50, and 100\% refers to the propagated ratio of the videos.}
\label{fig:additional_qualitative}
\end{figure*}

\end{document}